\documentclass[acmtog,nonacm]{acmart}

\renewcommand\footnotetextcopyrightpermission[1]{} 
\AtBeginDocument{%
  }

\acmSubmissionID{1064}

\begin{document}

\title{Parametric Gaussian Human Model: Generalizable Prior for Efficient and Realistic Human Avatar Modeling}


\author{Cheng Peng}
\authornote{Both authors contributed equally to this research.}
\affiliation{%
  \institution{Tsinghua University}
  \city{Beijing}
  \country{China}
}

\author{Jingxiang Sun}
\authornotemark[1]
\affiliation{%
  \institution{Tsinghua University}
  \city{Beijing}
  \country{China}
}

\author{Yushuo Chen}
\affiliation{%
  \institution{Tsinghua University}
  \city{Beijing}
  \country{China}
}

\author{Zhaoqi Su}
\affiliation{%
  \institution{Tsinghua University}
  \city{Beijing}
  \country{China}
}

\author{Zhuo Su}
\affiliation{%
  \institution{ByteDance}
  \city{Shanghai}
  \country{China}
}

\author{Yebin Liu}
\authornote{Corresponding author}
\affiliation{%
  \institution{Tsinghua University}
  \city{Beijing}
  \country{China}
}

\renewcommand{\shortauthors}{Peng et al.}

\begin{abstract}
Photorealistic and animatable human avatars are a key enabler for virtual/augmented reality, telepresence, and digital entertainment. While recent advances in 3D Gaussian Splatting (3DGS) have greatly improved rendering quality and efficiency, existing methods still face fundamental challenges, including time-consuming per-subject optimization and poor generalization under sparse monocular inputs. In this work, we present the \textit{Parametric Gaussian Human Model (PGHM)}, a generalizable and efficient framework that integrates human priors into 3DGS for fast and high-fidelity avatar reconstruction from monocular videos. \textit{PGHM} introduces two core components: (1) a \textit{UV-aligned latent identity map} that compactly encodes subject-specific geometry and appearance into a learnable feature tensor; and (2) a \textit{disentangled Multi-Head U-Net} that predicts Gaussian attributes by decomposing static, pose-dependent, and view-dependent components via conditioned decoders. This design enables robust rendering quality under challenging poses and viewpoints, while allowing efficient subject adaptation without requiring multi-view capture or long optimization time. Experiments show that \textit{PGHM} is significantly more efficient than optimization-from-scratch methods, requiring only approximately 20 minutes per subject to produce avatars with comparable visual quality, thereby demonstrating its practical applicability for real-world monocular avatar creation.
\end{abstract}


\begin{teaserfigure}
  \includegraphics[width=\linewidth]{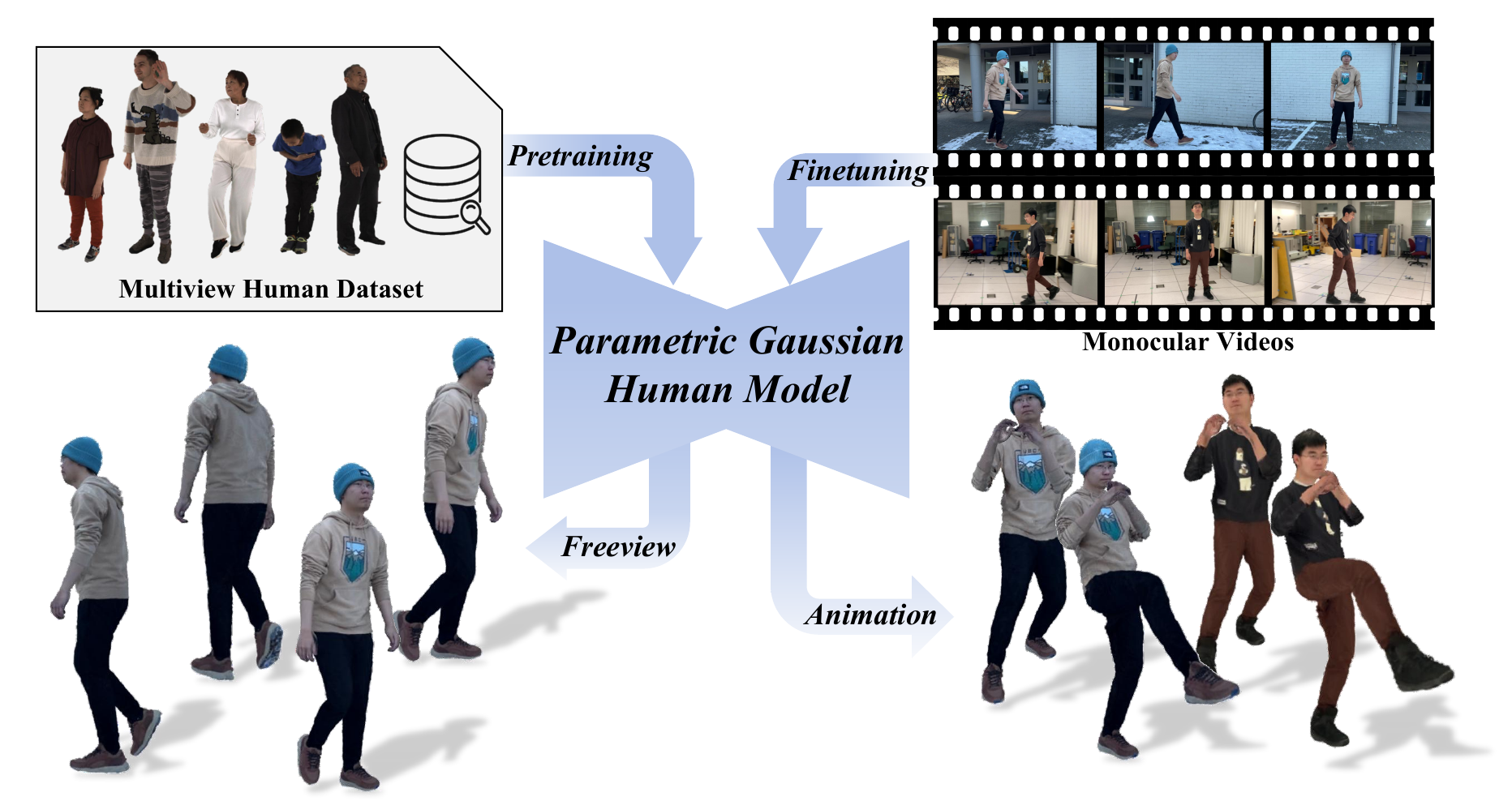}
  \vspace{-5mm}
  \caption{We introduce the Parametric Gaussian Human Model (PGHM), a generalizable prior for efficient and realistic human avatar modeling. After being trained on a large-scale, high-quality multiview human dataset, PGHM can be efficiently fine-tuned using monocular single-person videos. This enables accurate avatar reconstruction and supports both free-viewpoint rendering and animation.
}
  \Description{We introduce the Parametric Gaussian Human Model (PGHM), a generalizable prior for efficient and realistic human avatar modeling.}
  \label{fig:teaser}
\end{teaserfigure}

\received{20 February 2007}
\received[revised]{12 March 2009}
\received[accepted]{5 June 2009}

\maketitle

\section{Introduction}

Photorealistic and animatable human avatars represent a crucial research direction in 2D and 3D vision, enabling applications in virtual/augmented reality, telepresence, digital entertainment, etc.. 
Traditional mesh-based or point-based human avatars suffer from fundamental limitations - predefined topologies and unstructured representations, respectively - that hinder realistic avatar creation. In recent years, while the emergence of NeRF has significantly advanced human avatar quality, NeRF's intrinsic slow rendering and expensive optimization remain fundamental bottlenecks.
  
Recently, the explicit 3D Gaussian Splatting representation has emerged as a breakthrough technology that combines accelerated rendering speeds with superior visual quality, thereby significantly benefiting photorealistic human avatar creation. 
However, current Gaussian avatar approaches still face fundamental challenges. On one hand, human avatar from multi-view video inputs, e.g., Animatable Gaussians~\cite{li2024animatablegaussians}, achieve high-quality rendering results and pose-dependent dynamics by introducing pose-dependent Gaussian maps, yet require 1-2 days of training per subject. Besides, multi-view inputs require complicated data capture setups.
On the other hand, avatars from monocular inputs~\cite{hu2024gaussianavatar,moon2024exavatar} combine 3DGS with SMPL-UV or SMPL-X geometric models to learn the pose-dependent effects, yet suffer from blurriness and lack appearance details, as monocular inputs 
tend to struggle with generalizing to diverse or unseen poses due to incomplete observations and inherent limitations in the training data.
Therefore, to achieve efficient and high-quality human avatars from monocular inputs, the key lies in developing a generalizable parametric model that can learn human priors from large-scale data while maintaining the representational advantages of 3D Gaussians.

We argue that incorporating parametric human priors into the Gaussian-based human avatar is essential, as it enables rapid subject-specific adaptation through the learned prior, and more robust performance under challenging input conditions.
In this paper, we propose Parametric Gaussian Human Model, which learns generalizable avatar priors from large-scale data while enabling fast adaptation to novel subjects. Our Parametric Gaussian Human Model achieves this through two key designs. 
Firstly, we introduce the UV-Aligned Latent Identity Map, which encodes identity-specific attributes such as facial features and clothing geometry into a compact, learnable feature tensor. This approach differs from the GaussianAvatar's~\cite{hu2024gaussianavatar} reliance solely on UV position maps for pose information. Our design enhances the original framework by utilizing the UV-Aligned Latent Identity Map as a control signal for identity. During the fine-tuning phase, we can optimize this map alone to effectively capture personalized characteristics. This method contributes to rapid fine-tuning and improved identity control.
Secondly, for better learning Gaussian attributes from the Latent Identity Map, we propose the disentangled Multi-Head U-Net that explicitly models static, dynamic, and view-dependent effects through pose/view-conditioned decoders, achieving consistent performance across unseen identities, diverse poses and challenging viewpoints. Together, we achieve robust and high-fidelity human avatars through parametric Gaussian priors learned from large human datasets combined of selected MVHumanNet~\cite{xiong2024mvhumannet} and DNA-Rendering\cite{cheng2023dna} datasets, enabling fast adaptation for a personalized avatar from a monocular video. Compared with the concurrent work Vid2Avatar-Pro~\cite{guo2025vid2avatarpro}, our method achieves comparable optimization time for monocular-input personalized avatars, while Vid2Avatar-Pro requires an additional 36–48 hours for per-identity mesh template reconstruction.
Experiments demonstrate that our method achieves SOTA avatar rendering quality and avatar training efficiency. The contributions of our paper are summarized as follows:

\begin{itemize}
    \item We propose the Parametric Gaussian Human Model, a framework that integrates parametric human priors with 3D Gaussian Splatting, enabling fast and high-fidelity avatar adaptation to novel subjects.
    \item A UV-aligned latent identity map that compactly encodes subject-specific attributes into a learnable feature tensor, allowing memory-efficient personalization while preserving fine details.
    \item A disentangled Multi-Head U-Net architecture that dynamically decomposes Gaussian properties into static, pose-dependent, and view-aware components through conditioned decoders, ensuring dynamic details under challenging poses and viewpoints.
\end{itemize}
\section{Related Work}

\subsection{3D Human Avatar Modeling}

\paragraph{Avatar Modeling from Monocular Video.}
The advent of neural radiance fields (NeRF)~\cite{mildenhall2020nerf} has spurred a wave of \textit{implicit} avatar reconstruction approaches that fit articulated NeRFs to monocular video\cite{weng2022humannerf, jiang2022selfrecon, jiang2022neuman, su2021a-nerf, su2022danbo, chen2021animatable, su2023npc, Feng2022scarf, te2022neural}. Given the inherent noise in monocular pose estimation, many of these methods jointly refine motion trajectories during inverse rendering~\cite{weng2022humannerf, jiang2022neuman, guo2023vid2avatar, jiang2023instantavatar, yu2023monohuman}.
Although effective, the learned deformation fields often overfit to training sequences, resulting in artifacts under novel or out-of-distribution poses.

Recent advances replace implicit volumetric representations with \textit{explicit} 3D Gaussian Splatting (3DGS), significantly improving rendering speed and simplifying optimization.
Some approaches optimize Gaussian parameters per subject~\cite{Lei_2024_CVPR, shao2024splattingavatar, svitov2024hahahighlyarticulatedgaussian, li2024gaussianbodyclothedhumanreconstruction}, while others utilize neural networks to predict Gaussian attributes for more efficient personalization~\cite{qian20233dgsavatar, wen2024gomavatar, kocabas2024hugs, Hu_2024_CVPR, hu2024gaussianavatar, li2023human101, moon2024exavatar, liu2024gvareconstructingvivid3d}.
Despite the efficiency gains, current per-subject fittings often suffer from blurry textures and limited generalization to novel motions.
Our method addresses these limitations by pretraining a 3DGS human prior on a large corpus of dynamic human sequences, then adapting it to short monocular videos for sharper textures and improved motion robustness.

\paragraph{Avatar Modeling from Multi‑View Videos.}
Calibrated multi-camera systems enable high-fidelity avatar reconstruction by jointly modeling geometry and appearance across views~\cite{liu2021neural, bagautdinov2021driving, remelli2022drivable, peng2021neural, habermann2021, 10.1145/3478513.3480545, peng2021animatable, xu2022sanerf, HVTR:3DV2022, 2021narf, li2022tava, zhang2021stnerf, ARAH:ECCV:2022, chen2024meshavatar, saito2024rgca, zheng2023avatarrex, li2023posevocab, 10.1145/3697140, shen2023xavatar, yin2023hi4d, zheng2022structured}.
Early works often rely on canonical implicit fields to support non-rigid deformation, but they come at a high computational cost.
With the introduction of 3DGS~\cite{kerbl3Dgaussians}, recent methods replace volumetric rendering with Gaussian splatting and attach Gaussians to skeletal joints~\cite{li2024animatablegaussians, zielonka2023drivable3dgaussianavatars, moreau2024human, Pang_2024_CVPR, PhysAavatar24, jung2023deformable3dgaussiansplatting}.
To further enhance the representation capability for human dynamics and regularize surface reconstruction, 2D-map parameterizations are employed to guide geometry and texture learning~\cite{Pang_2024_CVPR, hu2024gaussianavatar, li2024animatablegaussians}.
Nevertheless, current designs are either manually engineered or limited to a fixed identity, restricting generalization and scalability.

\subsection{Avatar Prior Models}

\paragraph{Statistical and Regression Priors.}
Full-body statistical models such as SMPL and SMPL-X, along with part-specific 3DMMs for facial modeling, offer low-dimensional shape spaces governed by joint angles~\cite{loper2015smpl,SMPL-X:2019,Joo_2018_CVPR}. However, their minimalist templates lack garment wrinkles and fine-scale facial details. Pixel-aligned regressors—including PIFu, PIFuHD, ICON, ECON, SITH, ARCH/ARCH-H, and MIGS~\cite{saito2019pifu, saito2020pifuhd, xiu2022icon, xiu2023econ, ho2024sith, huang2020arch, He_2021_ICCV, chatziagapi2024migs}—predict high-resolution surface detail from single images, but they are trained on limited-scale static scan datasets. As a result, the reconstructed avatars are typically non-rigged or exhibit unnatural deformations during animation. Generalizable human rendering methods~\cite{kwon2021neural, kwon2024ghg, sun2024metacap, Zhao_2022_CVPR,Chen_2023_CVPR, chen2022gpnerf, zheng2024gpsgaussian, pan2024humansplat} synthesize novel views from sparse camera inputs but offer limited articulation control. In contrast, part-specific priors such as HeadGap, SEGA, URAvatar, URHand, OHTA, CAFCA, and Lisa~\cite{zheng2024headgap, guo2025segadrivable3dgaussian,
10.1145/3528223.3530143,
buhler2023preface, buehler2024cafca, li2024uravatar, Corona_2022_CVPR, Moon_2024_CVPR, chen2024urhand, zheng2024ohta} reconstruct highly detailed and realistically deformable heads or hands, yet they do not generalize to diverse full-body clothing. Our concurrent work~\cite{guo2023vid2avatar} introduces a unified clothed-human prior that enables high-fidelity reconstruction across varied identities and garments, with robust generalization to novel poses.

\paragraph{Generative Avatar Priors}
Recent progress in generative 3D avatar modeling can be broadly categorized into GAN-based and diffusion-based approaches. GAN-based methods have demonstrated strong capabilities in generating 3D-aware avatars from single-view image datasets~\cite{bergman2022generative, hong2022eva3d, dong2023ag3d, sun2023next3d, xu2024xagen, zhang2023getavatar, abdal2024gaussian}. GSM~\cite{abdal2024gaussian} proposes a surface-constrained Gaussian representation based on shell maps, which enhances training efficiency while preserving visual quality. On the diffusion side, latent diffusion models (LDMs) have enabled more flexible and semantically grounded 3D avatar generation. StructLDM~\cite{hu2025structldm} introduces a structured, 3D-aware LDM trained on a semantically aligned latent space to facilitate controllable generation and editing. Recently, IDOL~\cite{zhuang2024idolinstantphotorealistic3d} and LHM~\cite{qiu2025lhm} employ transformer-based architectures in a feed-forward manner with single image input, enabling fast and scalable 3D human avatar modeling, while limited in capturing dynamic deformation details.

\section{Parametric Gaussian Avatar Prior}
In this section, we present the Parametric Gaussian Avatar Prior. In contrast to person-specific training tasks, initializing and training Gaussian-based multi-human prior models pose distinct challenges. This section introduces the Gaussian human representation, the carefully designed Gaussian Parametric Human Model, and how to utilize our model for avatar creation when given an input monocular video.

\subsection{Preliminaries}
\label{sec:preliminaries}

3D Gaussian Splatting~\cite{kerbl3Dgaussians} is a 3D point-based representation for efficient and realistic rendering. It is represented by a set of 3D Gaussians, each of which is parameterized by its 3D center position $\boldsymbol{\mu} \in \mathbb{R}^3$, scale $\mathbf{s} \in \mathbb{R}^3$, rotation parameterized as a quaternion $\mathbf{q} \in \mathbb{R}^4$, color $\mathbf{c} \in \mathbb{R}^3$, and opacity $\sigma \in \mathbb{R}$, and distributed as:
\vspace{-1mm}
$$
f(\mathbf{x} \mid \boldsymbol{\mu}, \boldsymbol{\Sigma}) \propto \exp \left(-\frac{1}{2}(\mathbf{x}-\boldsymbol{\mu})^{\top} \boldsymbol{\Sigma}^{-1}(\mathbf{x}-\boldsymbol{\mu})\right),
$$
where the covariance matrix $\Sigma=\mathbf{RSS}^{\top} \mathbf{R}^{\top}$ is factorized into a scaling matrix $S$ and a rotation matrix $R$ given by the quaternions $q$ and scaling $\mathbf{s}$. 

To enhance the adaptability of Gaussian representations for generating and driving new viewpoints of digital humans, inspired by GaussianAvatar\cite{hu2024gaussianavatar} and ExAvatar\cite{moon2024exavatar}, we constrain all Gaussian assets to be isotropic. This is achieved by fixing the scale degree of freedom to 1, while setting the rotation to the identity matrix and the opacity to 1.

\subsection{UV-aligned Gaussian Human Representation}


To learn a Gaussian attribute representation that incorporates human semantics, we define a UV-position map to represent the posed SMPL-X mesh vertex positions and a UV-Gaussian map to encode the attributes of the generated Gaussian points and their correspondence to the initial SMPL-X mesh, instead of using an orthogonal projection position map as in Animatable Gaussians \cite{li2024animatablegaussians}. The use of UV maps enables the placement of a greater number of effective Gaussian points per unit area, thereby improving model efficiency. Furthermore, by allocating a larger proportion of the UV map to the facial region, our approach captures richer facial details.

\paragraph{UV Position Map.} We adjust the SMPL-X human model to a predefined neutral A-pose. Then, the mesh position attributes are unwrapped and mapped to their corresponding positions on a UV map, constructing a UV map representation of the Gaussian digital human. This strategy enables efficient 2D representation for storing and representing 3D position information. Finally, based on the initialized SMPL-X mesh information, we can obtain the UV map $\mathcal{U}_\theta \in \mathbb{R}^{L \times L \times 3}$ corresponding to the digital human in pose $\theta$.

\paragraph{UV Gaussian Map.}  Building upon the UV Position Map as the initial position map for the Gaussian points, we further predict the pixelwise offsets for position ($\delta_\mathrm{position}$), scale ($\delta_\mathrm{scale}$), and color ($\delta_\mathrm{color}$) through the model. Similar to the UV Position Map, we organize these attributes into corresponding UV Gaussian maps: the position offset map $\mathcal{U}_p \in \mathbb{R}^{L \times L \times 3}$, the scale map $\mathcal{U}_s \in \mathbb{R}^{L \times L \times 1}$, and the color map $\mathcal{U}_c \in \mathbb{R}^{L \times L \times 3}$. In this way, each Gaussian point is aligned with the UV points sampled from the mesh.

\subsection{Parametric Prior Model Training}
\label{sec:parametric}

\begin{figure*}[h]
    \centering
    \includegraphics[width=\linewidth]{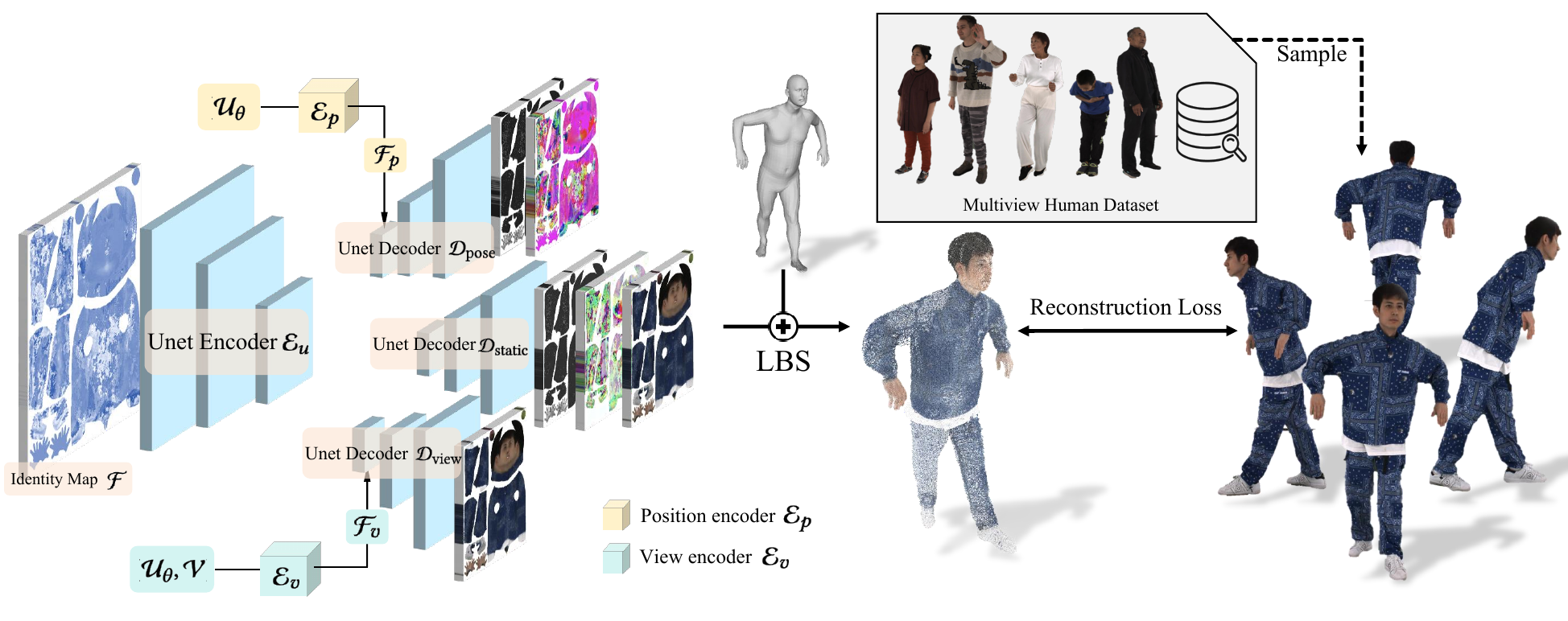}
    \caption{The overall pipeline of the parametric model training involves pre-training our model on a large-scale human dataset to obtain a robust human prior. This process consists of two key components: 1) a UV-aligned identity map to extract the appearance feature information of individuals, and 2) a Disentangled Multi-Head U-Net to decouple pose-dependent and view-dependent Gaussian attributes.}
    \vspace{-3mm}
    \label{fig:compare1}
\end{figure*}

To effectively learn a generalizable Gaussian avatar prior from diverse human subjects, we propose a unified parametric model that encodes large-scale human data into compact latent representations while preserving distinctive individual characteristics.
Specifically, we formalize our Gaussian avatar prior as a pipeline that first embeds each subject into a compact latent map, then expands this map through a lightweight decoder into a dense identity tensor, and finally feeds the tensor—together with pose and view cues—into a disentangled Multi-Head U-Net whose static, dynamic, and view branches jointly predict the canonical geometry/texture, pose-driven deformations, and view-dependent color of the UV-aligned Gaussian atlas. In the following, we provide detailed descriptions of each component.


\paragraph{UV-Aligned Latent Identity Map.} To parametrize the identity information of different characters (such as appearance, clothing, etc.), we employ a UV-aligned per-identity feature map to encode these attributes. The feature map contains n features $\mathcal{F}' \in \mathbb{R}^{C \times L' \times L'}$, where each feature represents the characteristic information of a distinct character identity. 
To maintain training efficiency and memory economy, we utilize a Feature Decoder to decode each feature into a tensor \( \mathcal{F} \in \mathbb{R}^{C \times L \times L} \), where \( L \) is twice the size of \( L' \).
This approach ensures that the feature maps remain consistent while containing richer information, which is beneficial for subsequent single-character fine-tuning.


\paragraph{Disentangled Multi-Head U-Net.}
For a specific character, there is a static geometric offset and static appearance relative to the SMPL-X shape. As the character moves, dynamic changes in geometry and appearance are introduced. Furthermore, varying viewpoints and lighting conditions induce view-dependent color shifts. Modeling such complex requirements therefore calls for a carefully designed architecture. To address this, we propose a MultiHead U-Net to predict the Gaussian properties of the character. This design enables the network to model static geometric and appearance offsets (\emph{w.r.t.} SMPL-X), dynamic deformations, as well as view-dependent appearance variations in a unified manner.

To accurately model the Gaussian representations, we designed a MultiHead U-Net architecture, which consists of a feature code encoder head \( \mathcal{E}_{unet} \) and three decoder heads \( \mathcal{D}_\text{static} \), \( \mathcal{D}_\text{pose} \), and \( \mathcal{D}_\text{view} \). The output from each layer of the encoder is connected in a U-shaped manner to all three decoders. For decoders \( \mathcal{D}_\text{pose} \) and \( \mathcal{D}_\text{view} \), we inject pose information and view information, respectively.

For the pose and view injection, we employ two lightweight convolutional encoders \( \mathcal{E}_{p} \) and \( \mathcal{E}_{v} \) to extract feature information. To obtain pose-dependent information, we use \( \mathcal{E}_{p} \) to encode the UV position map, yielding pose-dependent feature \( \mathcal{F}_{p} \):

$$\mathcal{F}_{p} = \mathcal{E}_{p}\left(\mathcal{U}_{\theta}\right)$$

For view-dependent information, we first construct a view direction map \( \mathcal{V} \) based on the normal map to model view-dependent variance, similar to NeRF-based approaches ~\cite{peng2022animatable}. We then use \( \mathcal{E}_{v} \) to extract the view features \( \mathcal{F}_{v} \):

$$
\mathcal{F}_{v} = \mathcal{E}_v\left(\left[\mathcal{U}_{\theta}, \mathcal{V}\right]\right)
$$

To extract the identity information of the subject, we employ the U-Net encoder \( \mathcal{E}_u \) to extract appearance features \( \mathcal{F}_{u} \) from the previously mentioned UV-Aligned Latent Identity Map \( \mathcal{F} \) :

\[
\mathcal{F}_{u} = \mathcal{E}_u\left(\mathcal{F}\right)
\]

After obtaining the three types of features described above (\( \mathcal{F}_u \), \( \mathcal{F}_p \), and \( \mathcal{F}_v \)), we further utilize the three decoder heads of the U-Net to extract Gaussian representation information.

For the first decoder head \( \mathcal{D}_{static} \), we predict the static Gaussian representations, including the Gaussian position UV map (\( \mathcal{U}_p \)), Gaussian scale UV map (\( \mathcal{U}_s \)), and Gaussian color UV map (\( \mathcal{U}_c \)), as follows:

\[
\mathcal{U}_p, \ \mathcal{U}_s, \ \mathcal{U}_c = \mathcal{D}_\text{static}\left(\mathcal{F}_u\right)
\]

For the second decoder head \( \mathcal{D}_\text{pose} \), we predict dynamic, pose-dependent Gaussian position offset UV map (\( \mathcal{U}_p^o \)) and Gaussian scale offset UV map (\( \mathcal{U}_s^o \)), conditioned on both the identity features \( \mathcal{F}_u \) and the pose features \( \mathcal{F}_p \):

\[
\mathcal{U}_p^o, \ \mathcal{U}_s^o = \mathcal{D}_\text{pose}\left(\mathcal{F}_u, \mathcal{F}_p\right)
\]

For the third decoder head \( \mathcal{D}_\text{view} \), we predict the Gaussian color offset UV map (\( \mathcal{U}_c^o \)), which is both pose and view-dependent, by conditioning on the identity features \( \mathcal{F}_u \) and the view features \( \mathcal{F}_v \):

\[
\mathcal{U}_c^o = \mathcal{D}_\text{view}\left(\mathcal{F}_u, \mathcal{F}_v\right)
\]


\paragraph{Deformation.}
After all, we obtain the Gaussian position UV map (\( \mathcal{U}_p \)), Gaussian scale UV map (\( \mathcal{U}_s \)), Gaussian color UV map (\( \mathcal{U}_c \)), Gaussian position offset UV map (\( \mathcal{U}_p^o \)), Gaussian scale offset UV map (\( \mathcal{U}_s^o \)), and Gaussian color offset UV map (\( \mathcal{U}_c^o \)).
Due to our UV map design, we can use these Gaussian UV maps to index various Gaussian attributes, including \( \mathbf{V}_p \), \( \mathbf{V}_s \), \( \mathbf{V}_c \), \( \mathbf{V}_p^o \), \( \mathbf{V}_s^o \), and \( \mathbf{V}_c^o \). Based on these attributes, we separately construct pose-independent (static) and pose-dependent (dynamic) Gaussian deformations.
For the pose-independent Gaussian deformation and the pose-dependent Gaussian deformation, we compute the canonical space pose-independent Gaussian locations $\mathbf{V}_\text{sta}$ and canonical space pose-dependent Gaussian locations $\mathbf{V}_\text{dyn}$ as follows:

\[
\overline{\mathbf{V}}_\text{sta} = \overline{\mathbf{V}} + \mathbf{V}_{p}
\]
\[
\overline{\mathbf{V}}_\text{dyn} = \overline{\mathbf{V}} + \mathbf{V}_{p} + \mathbf{V}_{p}^{o}
\]

The final vertex positions are then computed using Linear Blend Skinning (LBS) as:
\[
\mathbf{V}_\text{sta} = \operatorname{LBS}\left(\overline{\mathbf{V}}_\text{sta}, \theta\right)
\]
\[
\mathbf{V}_\text{dyn} = \operatorname{LBS}\left(\overline{\mathbf{V}}_\text{dyn}, \theta\right)
\]
where \( \overline{\mathbf{V}} \) denotes the canonical vertex positions, \( \mathbf{V}_{p} \) and \( \mathbf{V}_{p}^{o} \) are the pose-independent and pose-dependent Gaussian position offsets indexed from the corresponding UV maps, \( \theta \) denotes the pose parameters.

\paragraph{Rasterization.}To render the animated 3D geometry, we utilize the 3D Gaussian Splatting (3DGS) rendering pipeline [22], defined as follows:
\[
\mathbf{I_s} = f\left(\mathbf{V}_{sta}, \mathbf{V_c}, \mathbf{K}, \mathbf{E}\right),
\]
\[
\mathbf{I_d} = f\left(\mathbf{V}_{dyn}, \mathbf{V_c} + \mathbf{V_c}^o, \mathbf{K}, \mathbf{E}\right),
\]
where \( f \) denotes the 3DGS rendering function, while \( \mathbf{K} \) and \( \mathbf{E} \) are the camera intrinsic and extrinsic matrices, respectively. 

As previously mentioned, all Gaussian primitives are constrained to be isotropic to enhance generalization capability. Therefore, both the rotation and the opacity of each Gaussian are fixed to the identity and unity, respectively, and are omitted from the above equations for clarity.

\paragraph{Training objectives.}
During the training process of the parametric model, we jointly optimize the per-identity feature map, Feature Encoder, and Multihead-Unet modules. Additionally, since the pose annotations in the training data are not entirely accurate, we further optimize the human poses in the training data during the training process. To ensure the reliability of both pose-independent and pose-dependent information, we employ both types of rendered results as supervision. The final training objective is defined as follows:

\vspace{-2mm}
\begin{align*}
\mathcal{L} =\; & 
\lambda_1 \big[\, 
    \mathcal{L}_1\big(I_s, I_{gt}\big) + 
    \mathcal{L}_1\big(I_d, I_{gt}\big) 
\,\big] \\
& +\; 
\lambda_2 \big[\, 
    \mathcal{L}_{\mathrm{ssim}}\big(I_s, I_{gt}\big) + 
    \mathcal{L}_{\mathrm{ssim}}\big(I_d, I_{gt}\big) 
\,\big] \\
& +\; 
\lambda_3 \big[\, 
    \mathcal{L}_{\mathrm{lpips}}\big(I_s, I_{gt}\big) + 
    \mathcal{L}_{\mathrm{lpips}}\big(I_d, I_{gt}\big) 
\,\big] \\
& +\; 
\lambda_4\, \mathcal{L}_{\mathrm{reg}}
\end{align*}

In our loss function, the regularization term $\mathcal{L}_{\mathrm{reg}}$ is designed to prevent unreasonable scale and position values. Specifically, we apply an ($L_2$) penalty to both the scale offset and position offset parameters. This encourages the optimized scale and position to remain close to their initial values, thus avoiding implausible transformations during training.

\section{Personalization with Parametric Avatar Prior}

\begin{figure}[h]
    \centering
    \vspace{-4mm}
    \includegraphics[width=\linewidth]{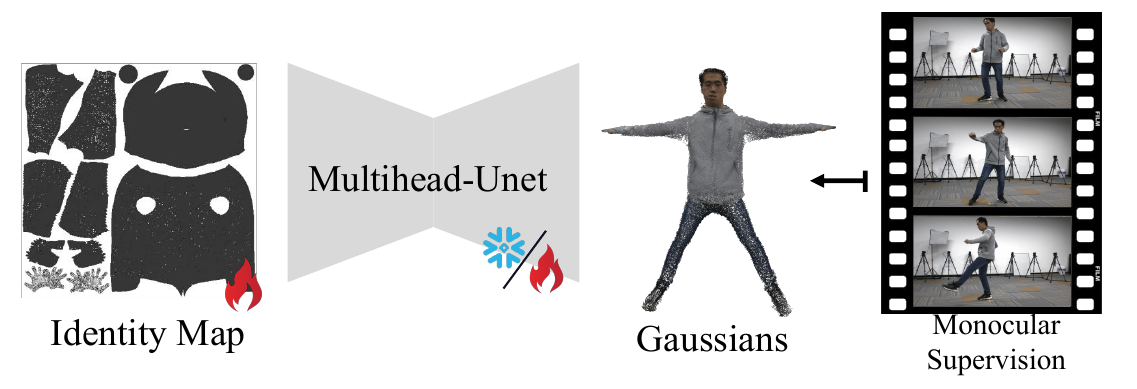}
    \vspace{-7mm}
    \caption{The personalization stage consists of two steps: 1) Identity Map adaptation and 2) Joint tuning of the Identity Map and Multi-Head U-Net.}
    \vspace{-4mm}
    \label{fig:finetune}
\end{figure}

As shown in Fig \ref{fig:finetune}, we adopt a simple two-stage strategy to personalize our parametric avatar prior to a specific subject from a monocular video.
\paragraph{Identity Map adaptation.} As discussed in Section \ref{sec:parametric}, we usually expand discrete identity map 
$\mathcal{F}$ from a smaller feature $\mathcal{F}'$ using a feature encoder. During personalization, we skip the encoder and instead initialize a new learnable feature tensor to directly replace the expanded map. This approach makes the tuning process more efficient and helps retain subject-specific details that might otherwise be lost through the encoder.
\paragraph{Joint tuning of Identity Map and Multi-Head U-Net.} Because monocular videos provide only limited supervision, adapting the feature code alone isn’t enough. To better capture subject-specific appearance and geometry, we further fine-tune both the feature code and the Multi-Head U-Net together. This joint optimization allows the model to better generalize to novel inputs while keeping the number of trainable parameters small.

\vspace{-2mm}
\section{Experiment}

\subsection{Dataset}

We trained our prior model using the MVHumanNet~\cite{xiong2024mvhumannet} and DNA-Rendering~\cite{cheng2023dna} datasets, both of which are large-scale multi-view human datasets comprising thousands of samples. Given the limited availability of accurate pose annotations in DNA-Rendering, we selectively fit poses for 300 samples from parts 3 to 6 of this dataset. To further increase the diversity of identities, we additionally sampled 300 identities from MVHumanNet, resulting in a mixed training set. It is worth noting that the DNA-Rendering dataset exhibits superior temporal continuity, whereas MVHumanNet consists of frames sparsely sampled from original videos, leading to lower temporal consistency.



For our evaluation, we conducted both qualitative and quantitative experiments on the NeuMan~\cite{jiang2022neuman} and THuman4.0~\cite{zheng2022structured} datasets. The NeuMan dataset consists of monocular video sequences captured in natural environments, featuring human subjects walking through various settings. We followed the methodologies outlined in ~\cite{jiang2022neuman} for splitting the training and test sets, with pose initialization based on ~\cite{moon2024exavatar}. In contrast, the THuman4.0 dataset is a high-resolution multi-view collection (1330 × 1150 pixels) known for its richly textured and dynamically detailed subjects. Here, we selected 500 frames as the training set and the subsequent 50 frames for testing, utilizing the dataset's original poses for initialization. Our analyses of these two datasets thoroughly assess the effectiveness of our learned priors.

\subsection{Experiment Settings}

During the prior training phase, our model was trained on a combined dataset consisting of MVHumanNet and DNA-Rendering samples. We utilized a total batch size of 32, distributed across eight Nvidia V100 GPUs, for a total of 50k iterations. This extensive training procedure lasted approximately seven days. The loss function was composed of several terms with their respective weights: L1 with $\lambda_1=0.8$, SSIM with $\lambda_2=0.2$, LPIPS with $\lambda_3=0.2$, and regularization set to $\lambda_4=0.1$.

For the reconstruction process, we typically execute 1k iterations dedicated to feature map alignment, followed by another 1k iterations for model fine-tuning on each sequence from the NeuMan dataset. This entire procedure requires around 20 minutes per sequence when run on a single Nvidia V100 GPU. 

\subsection{Comparison}


\begin{table}[h!]
\centering
\vspace{-2mm}
\begin{tabular}{lcccc}
\hline
Method & PSNR $\uparrow$ & SSIM $\uparrow$ & LPIPS $\downarrow$ & training time\\
\hline
HumanNeRF & 27.06 & 0.967 & 0.0252 & 72 hours\\
InstantAvatar & 28.74 & 0.972 & 0.0277 & 5 mins\\
GaussianAvatar & 28.90 & 0.969 & 0.0242 & 6 hours\\
NeuMan & 29.32 & 0.972 & 0.0201 & 7 days\\
ExAvatar & 31.42 & 0.981 & 0.0190 & 5 hours\\
\hline
Ours & $\mathbf{31.85}$ & $\mathbf{0.987}$ & $\mathbf{0.0171}$ & 20min\\
\hline
\end{tabular}
\caption{\textbf{Quantitative comparisons on Neuman test dataset.} We outperform other methods and achieve a significant improvement in training efficiency compared to other Gaussian-based approaches.}
\label{tab:neuman_comparison} 
\vspace{-6mm} 
\end{table}

\begin{table}[h!]
\centering
\begin{tabular}{lcccc}
\hline
Method & PSNR $\uparrow$ & SSIM $\uparrow$ & LPIPS $\downarrow$ \\
\hline
GaussianAvatar & 25.96 & 0.9682 & 0.0242 \\
ExAvatar & 27.87 & 0.9738 & 0.0251 \\
\hline
Ours & $\mathbf{28.23}$ & $\mathbf{0.9771}$ & $\mathbf{0.0184}$ \\
\hline
\end{tabular}
\caption{\textbf{Quantitative comparisons on Thuman test dataset result.} We outperform other methods on Thuman, a human dataset with rich textures and dynamic details.}
\label{tab:thuman_comparison} 
\vspace{-7mm} 
\end{table}

\begin{figure}[h]
    \centering
    \includegraphics[width=\linewidth]{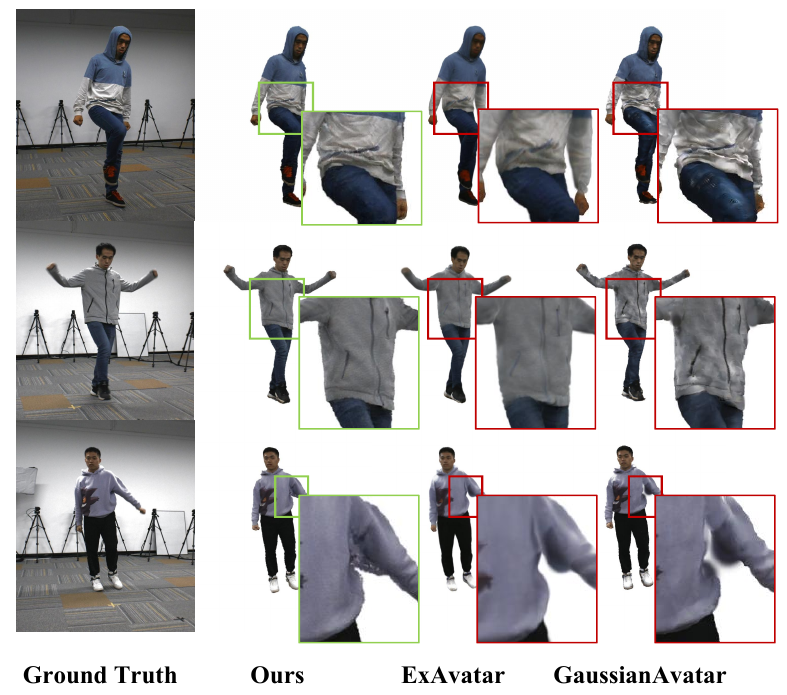}
    \vspace{-7mm}
    \caption{\textbf{Qualitative comparisons on Thuman test dataset.}Our method outperforms other approaches in both geometric details and color appearance.}
    \vspace{-5mm}
    \label{fig:thuman_compare}
\end{figure}

\begin{figure}[h]
  \centering
  \vspace{-5mm}
  \includegraphics[width=\linewidth]{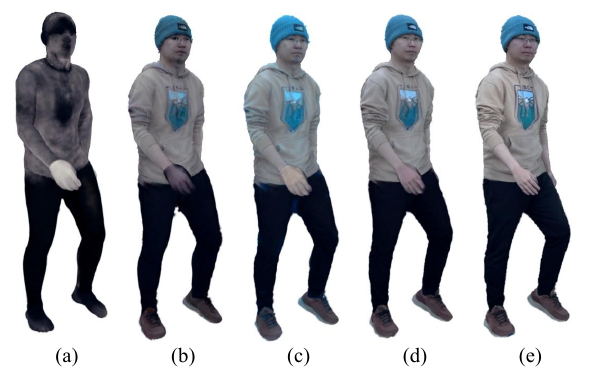}
  \vspace{-8mm}
  \caption{Ablation on ID number effect on feature map tuning stage: (a)-(d) 0, 6, 60 and 600 IDs pretrained (e) Jointly finetune both feature map and unet}
  \label{fig:ablation1}
  \vspace{-4mm}
\end{figure}



We mainly compared our method with GaussianAvatar~\cite{hu2024gaussianavatar} and ExAvatar~\cite{moon2024exavatar}, both of which are 3DGS-based approaches for constructing human avatars. To quantitatively evaluate performance, we employed established image quality metrics, including Peak Signal-to-Noise Ratio (PSNR), Structural Similarity Index Measure (SSIM)~\cite{wang2004image}, and Learned Perceptual Image Patch Similarity (LPIPS)~\cite{zhang2018the}. All metrics were computed over the entire image, with backgrounds set to white.

We present the quantitative evaluation results on two benchmark datasets. As summarized in Tab \ref{tab:neuman_comparison} and Tab \ref{tab:thuman_comparison}, our proposed method consistently surpasses all existing baselines across a comprehensive set of metrics, demonstrating superior capability in recovering fine-grained dynamic appearance details and more plausible cloth movement.

\begin{figure*}[h]
    \centering
    \includegraphics[width=\linewidth]{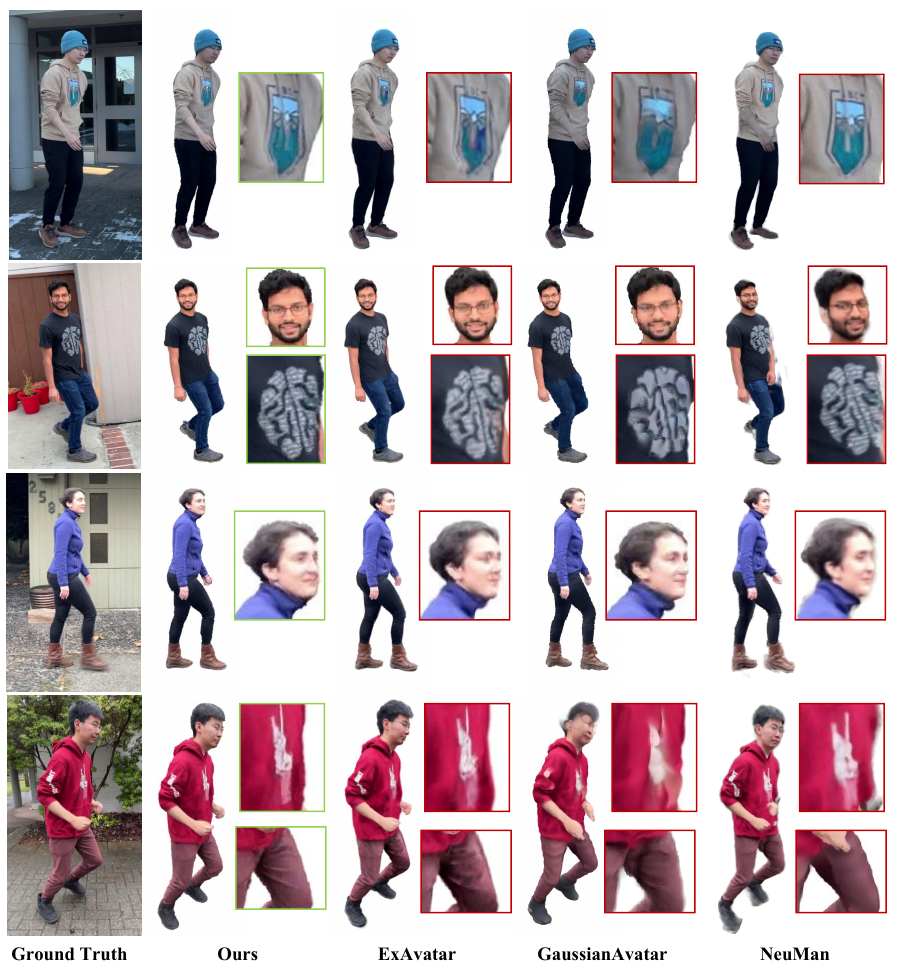}
    \caption{Qualitative comparisons on NeuMan test dataset. Our method outperforms other approaches in both geometric details and color appearance.}
    \label{fig:neuman_compare}
\end{figure*}

Qualitative comparisons, illustrated in Figures \ref{fig:neuman_compare} and \ref{fig:thuman_compare}, demonstrate the advantages of our approach over baseline methods, producing reconstructions with significantly enhanced detail fidelity and accurate pose-dependent dynamic texture information. On the NeuMan dataset, ExAvatar maintains a reasonable level of detail but struggles on the THuman4.0 dataset, where its performance is limited by complex regularization constraints that oversmooth results and blur fine details. In contrast, the GaussianAvatar method fails to effectively disentangle lighting and motion, resulting in unrealistic lighting artifacts. Our approach addresses these limitations through two key innovations: the introduction of learned priors to reduce reliance on heavy regularization, thus preserving intricate details, and the implementation of a multi-head U-Net architecture that effectively disentangles pose and illumination information, leading to more accurate and expressive reconstructions.


\subsection{Ablation Study}

To further validate the effectiveness of our proposed priors and the multi-head U-Net architecture, we conducted a series of ablation experiments. These experiments are designed to systematically analyze the contributions of each component to the overall performance of our framework.

\paragraph{Effectiveness of the Learned Priors}
Figure \ref{fig:ablation1} presents a comprehensive analysis of the impact of our learned priors. Subfigures (a)–(d) correspond to models fine-tuned on identity maps pre-trained with increasing numbers of identities: specifically, 0, 6, 60, and 600 IDs, respectively. Subfigure (e) shows the results after jointly fine-tuning both the identity map and the multi-head U-Net.

The results clearly demonstrate that scaling up the number of identities used during prior training significantly enhances the representational capacity of the identity map. As the number of pre-trained identities increases, the identity map exhibits greater generalization ability across unseen individuals and achieves faster convergence during fine-tuning. Moreover, even when only the identity map is fine-tuned, the reconstructed geometry and texture details become noticeably richer and more plausible, highlighting the benefit of our prior learning strategy.

However, due to the inherent limitations in the diversity and size of available training data, further fine-tuning of both the identity map and the multi-head U-Net is necessary to achieve higher fidelity in both texture and geometry. Importantly, we observe that our two-stage fine-tuning approach dramatically improves training efficiency: compared to training the model from scratch, our method achieves superior results using an order of magnitude fewer optimization steps (e.g., 2,000 steps versus 20,000 steps).

\begin{figure}[h]
  \centering
  \vspace{-2mm}
  \includegraphics[width=\linewidth]{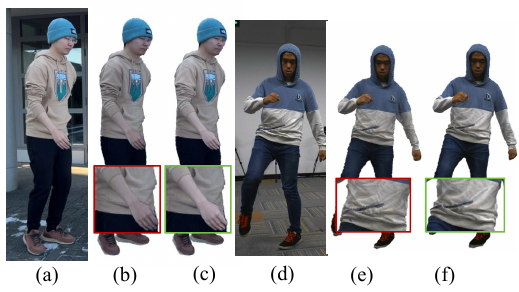}
  \vspace{-8mm}
  \caption{Ablation on multihead-unet: (a) (d) Ground Truth; (b) (e)Single-head U-Net Result; (c) (f) Multi-head U-Net Result.}
  \Description{ablation2}
  \label{fig:ablation2}
  \vspace{-5mm}
\end{figure}

\paragraph{Effectiveness of the Multi-Head U-Net Architecture}
Figure \ref{fig:ablation2} evaluates the efficacy of our multi-head U-Net design in disentangling pose and illumination effects. In each set of results, subfigures (a) and (d) depict the ground truth images; (b) and (e) show the outputs of a single-head U-Net; while (c) and (f) present results from our multi-head U-Net model.



The comparison demonstrates that our multi-head U-Net significantly enhances both the realism and accuracy of synthesized outputs. Specifically, disentangling appearance and lighting yields more plausible illumination in (c), while finer dynamic details such as wrinkles and folds in clothing are better preserved in (f), closely resembling the ground truth. In contrast, the single-head U-Net results (b, e) are noticeably darker and lack critical dynamic texture details, highlighting its limitations in modeling lighting and fine appearance. These results underscore the superiority of our multi-head U-Net architecture in generating detailed, physically consistent reconstructions.

Overall, the ablation studies demonstrate that both the learned priors and the multi-head U-Net architecture play critical roles in the success of our approach. The learned priors enhance identity generalization and accelerate convergence, while the multi-head U-Net enables effective disentanglement of pose and lighting, leading to more realistic and detailed human reconstructions.


\section{Conculsion}

We present \textit{Parametric Gaussian Human Model (PGHM)}, a novel framework that integrates parametric human priors into 3D Gaussian Splatting for efficient and high-fidelity monocular human avatar reconstruction. By introducing a UV-aligned latent identity map and a disentangled Multi-Head U-Net, PGHM enables fast subject-specific adaptation and robust rendering under challenging poses and viewpoints. Compared to existing methods, our approach achieves efficient training with only $\sim$20 minutes per subject, while maintaining competitive visual quality. We believe our method paves the way for scalable, efficient human avatar generation in immersive applications.

Our method currently presents two main limitations. First, it depends on input video sequences for optimizing the identity map, which restricts its use in cases with limited subject data. Future work will explore end-to-end feed-forward architectures that can infer identity features directly from fewer images, reducing per-subject optimization requirements.Second, the approach is less effective for subjects wearing loose or flowing clothing, like skirts or robes, due to challenges in modeling large non-rigid deformations. Addressing these issues will make our system more robust and widely applicable.

\bibliographystyle{ACM-Reference-Format}
\bibliography{reference}

\end{document}